\documentclass[journal]{IEEEtran}
\usepackage[colorlinks, linkcolor=blue, anchorcolor=blue, citecolor=blue]{hyperref}
\usepackage{algpseudocode} 
\usepackage{amsmath}
\usepackage{graphicx}
\usepackage{color}
\usepackage[numbers,sort&compress]{natbib}
\usepackage{amsmath,latexsym,amssymb,amsthm,array,amsfonts,algorithm,algpseudocode,booktabs,graphicx,subfigure,multirow,cuted,stfloats}

\ifCLASSINFOpdf
\else
\fi
\begin{document}
%
\title{A Survey on Computationally Efficient\\
       Neural Architecture Search}


%
%
%

\author{Shiqing Liu,
        Haoyu Zhang,
        and~Yaochu Jin,~\IEEEmembership{Fellow,~IEEE}
\thanks{S. Liu and Y. Jin are with the Chair of Nature Inspired Computing and Engineering, Faculty of Technology, Bielefeld University, 33619 Bielefeld, Germany. Y. Jin is also with the Department of Computer Science, University of Surrey, Guildford, GU2 7XH, United Kingdom. Email: yaochu.jin@uni-bielefeld.de.}
\thanks{H. Zhang is with the Engineering Research Center of Digitized Textile \& Apparel Technology, Ministry of Education, College of Information Science and Technology, Donghua University, Shanghai 201620, China.}
}

\maketitle

\begin{abstract}
Neural architecture search (NAS) has become increasingly popular in the deep learning community recently, mainly because it can provide an opportunity to allow interested users without rich expertise to benefit from the success of deep neural networks (DNNs). However, NAS is still laborious and time-consuming because a large number of performance estimations are required during the search process of NAS, and training DNNs is computationally intensive. To solve this major limitation of NAS, improving the computational efficiency is essential in the design of NAS. \textcolor{black}{However, a systematic overview of computationally efficient NAS (CE-NAS) methods still lacks. To fill this gap, we provide a comprehensive survey of the state-of-the-art on CE-NAS by categorizing the existing work into proxy-based and surrogate-assisted NAS methods, together with a thorough discussion of their design principles and a quantitative comparison of their performances and computational complexities.} The remaining challenges and open research questions are also discussed, and promising research topics in this emerging field are suggested.
\end{abstract}


%
\IEEEpeerreviewmaketitle

\section{Introduction}

Deep learning has played an important role in the area of machine learning. Currently, deep learning has been successfully applied to computer vision, including image classification \cite{he2016deep, huang2017densely, krizhevsky2012imagenet, simonyan2014very, szegedy2015going}, object detection \cite{girshick2015fast, ren2015faster, liu2016ssd, redmon2016you}, boundary detection \cite{xie2015holistically}, semantic segmentation \cite{chen2017deeplab, he2017mask, long2015fully}, pose estimation \cite{toshev2014human}, among many others. The technical design of deep learning heavily relies on DNNs because it automates the feature engineering process. The architectures of DNNs are usually developed for specific tasks and the associated weights can be obtained by a learning process. Only both are optimal at the same time can DNNs achieve promising performance. However, manually designing promising architectures of DNNs is a tedious task, mainly because it usually requires rich expertise in both deep learning and the investigated problems, and tries a great number of different hyperparameters to long-time tuning. These manually designed task-specific networks can not generalize to various application areas. For example, a network architecture designed for the image classification task may obtain inferior performance in object detection tasks. Moreover, it is sometimes needed to design network architectures under a limited computational budget (latency, memory, FLOPs, etc.) for different deployment scenarios. Handcrafted network architecture design is often inefficient to explore a large number of possibilities.

Recently, NAS appeared as a practical tool that allows engineers and researchers without expertise in deep learning to benefit from the success of DNNs. NAS focuses on searching effective task-specific network architectures on given datasets in an automatic manner. By defining a search space, which contains a large set of possible candidate network architectures, NAS can adopt the search strategies to explore extensive neural network architectures that have never been designed before. Most recently, compared with manually designed networks, NAS has obtained remarkable performance gain on representative benchmark datasets such as CIFAR10 \cite{krizhevsky2009learning}, CIFAR100 \cite{krizhevsky2009learning}, ImageNet \cite{deng2009imagenet}, etc., in terms of accuracy, model size, and computational complexity \cite{tan2019mnasnet, dong2019searching, zhang2020one, cai2018proxylessnas, lu2020neural}.

The purpose of NAS is to search for a network that minimizes some performance measures, such as error rates on unseen data (validation dataset). To guide the search process of NAS, early work \cite{baker2016designing, zoph2018learning, real2017large, real2019aging} usually adopt the simplest way to evaluate the performance of network candidates. A large number of networks are sampled from the search space and trained on training data from scratch, before their performance is evaluated on the validation dataset. Since training DNNs is itself computationally expensive, early NAS methods suffer from a high computational burden. For example, Zoph et al. \cite{baker2016designing} use reinforcement learning (RL) to design neural networks on the CIFAR10 dataset, which consumes 28 days on 800 high-performance graphics process units (GPU) cards. Unfortunately, not every researcher has access to sufficient computing resources. \textcolor{black}{Considerable computational overhead in evaluating network performance has been the bottleneck in the real-world application of NAS.} To address the above issue, researchers have made efforts on speeding up performance estimation. Consequently, the search time has been reduced from many GPU days down to several GPU hours. For example, recent work \cite{xu2021partially} allows a promising network architecture that can be searched within 0.1 GPU days on CIFAR10. 

\textcolor{black}{Little work have been reported to systematically review this emerging field. Therefore, this paper focuses on providing an overview of the research on improving the search efficiency of NAS methodologies.}  Based on whether the weights of network candidates are required when these network candidates are evaluated, they can be divided into two different categories, i.e., evaluating network candidates under the proxy metrics and surrogate-assisted neural architecture search. For proxy-based methods, the metrics still need the weights of architectures to be provided, which may introduce additional computational cost. \textcolor{black}{In contrast to the proxy-based methods, the performance evaluation of surrogate-assisted NAS methods only depends on the architecture of the network itself. Surrogate-based NAS methods usually rely on surrogates (also called performance predictor) to predict the performance of candidate networks, thereby avoiding the additional computational overhead. Training surrogates efficiently remains a challenging topic in NAS.}

\textcolor{black}{The rest of this paper is organized as follows. Section II provides the definition and mathematical formulation of NAS as an optimization problem, along with a brief overview of the development of NAS methods. Section III gives a detailed investigation of proxy-based NAS, which covers low-fidelity estimation, one-shot NAS and network morphism. Section IV presents a systematical analysis of existing surrogate-assisted NAS methods, including Bayesian optimization based methods, surrogate-assisted evolutionary based algorithms, federated NAS, and multi-objective NAS.  Section V summarizes the existing challenges and provides some insights into the future directions.}


\section{NAS}

NAS aims to search task-specific neural network architectures with high performance for a target dataset $D$ = $\{D_{tra}, D_{val}, D_{test}\}$ and releases engineers from the tremendous tedious network architecture designing process. NAS process can be modeled by a bilevel optimization problem, which can be formulated as follows:
\begin{align}\label{1}
\mathop W\nolimits_A^*  = \mathop {\arg \min }\limits_{A \in S} \mathop \mathcal{L}\nolimits_{tra} (N(A,\mathop W\nolimits_A ), D_{tra}),
\end{align}
  
\begin{align}\label{2}
\mathop A\nolimits^*  = \mathop {\arg \min }\limits_W \mathop \mathcal{L}\nolimits_{val} (N(A,\mathop W\nolimits_A^*), D_{val}).
\end{align}

In general, $S$ denotes the search space of the network architectures. $N(A, W_A)$ denotes the candidate architecture in the search space, where $W_A$ denotes the parameters associated to the network A. The goal of NAS is to search the network $A \in S$ that can achieve the promising performance on the validation set $D_{val}$ via minimizes the validation loss $\mathcal{L}_{val}$ according to Equation (2), and the parameters $W_A^*$ can be obtained through training model A on the training set $D_{tra}$ via minimizing the loss function $\mathcal{L}_{tra}$ according to Equation (1). 

\begin{figure}[H]
\centering 
\includegraphics[width=0.45\textwidth]{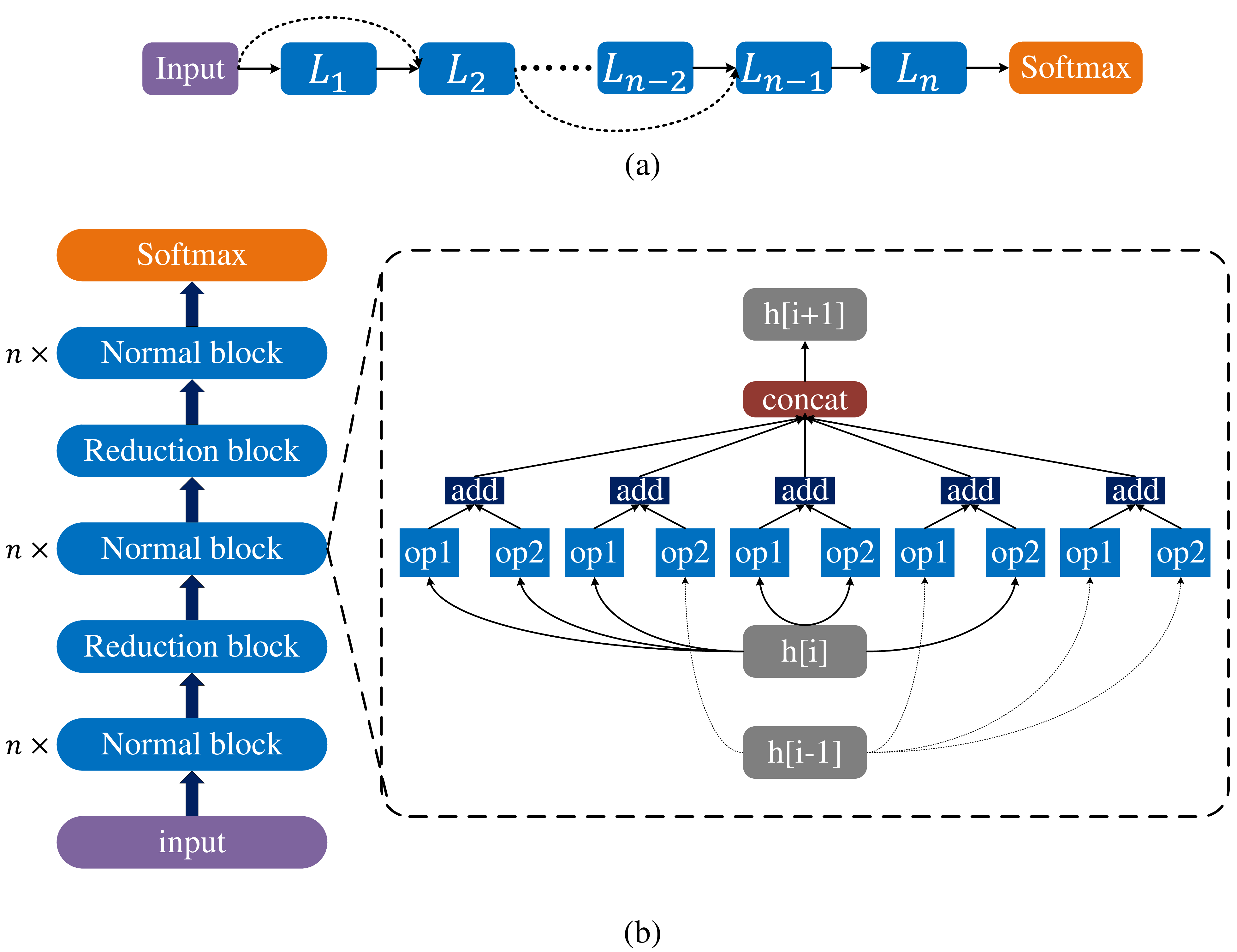} 
\caption{(a) An example of a network architecture represented in a macro search space. (b) An example of a network architecture represented in a micro search space. A typical example of a normal block structure is shown in the dashed box. Each block obtains the outputs from the previous block h[$i$] and previous–previous block h[$i-1$] as its inputs. The outputs of h[$i$] and h[$i-1$] are connected to operations (denoted as “op”). } 
\label{Fig_1}
\end{figure}

NAS search space is used to collect all possible candidate network architectures. Hence, the search space of NAS has a profound influence on the search efficiency and the performance of the designed models. In general, the search space can be divided into macro search space and micro search space. As shown in Fig.\ref{Fig_1}(a), the macro search space is proposed in the algorithm \cite{baker2016designing} by Google, which is over the entire network architecture, such as the number of layers n, the link manners for connections (e.g. shortcut \cite{he2016deep}), operation types (e.g. convolution and pooling), among others. As shown in Fig.\ref{Fig_1}(b), the micro search space is proposed in the algorithm \cite{zoph2018learning}, only covers repeated blocks in the whole network architecture. These blocks are constructed by complex multi-branch operations.

In theory, NAS can be seen as a complex optimization problem, which faces multiple challenges such as multiple conflicting optimization objectives, complex constraints, bi-level structures, expensive computational properties, among others. Early NAS research relies on evolutionary algorithms (EAs) and reinforcement learning (RL) to design the optimal network architecture for the given data set.

\subsection{RL-based NAS}

\begin{figure}[H]
\centering 
\includegraphics[width=0.45\textwidth]{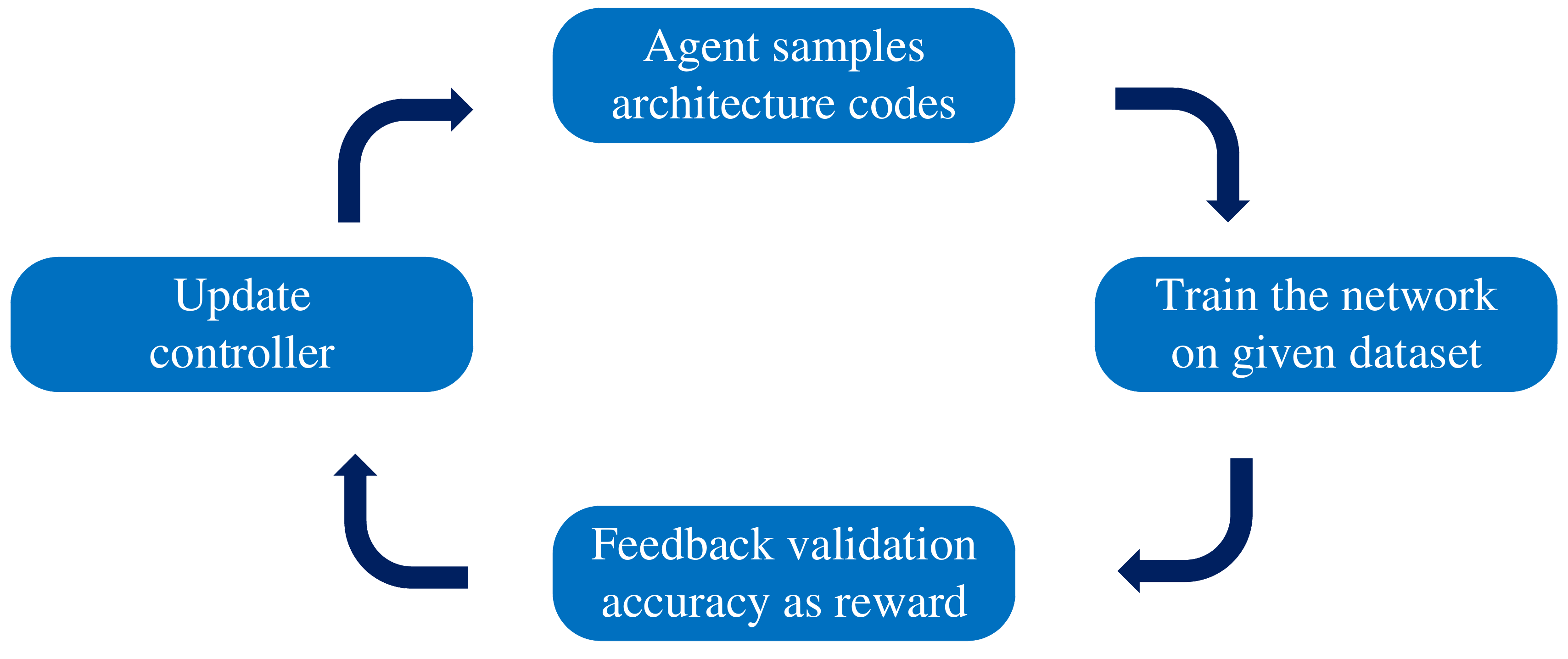} 
\caption{An overall framework of RL-based NAS algorithms.} 
\label{Fig_2}
\end{figure}

RL-based NAS methods consider the process of the design of network architecture as an agent’s action and train a meta-controller as a navigating tool to guide the search process. The overall framework of RL-based NAS algorithm is shown in Fig.\ref{Fig_2}. More specifically, A new candidate network is sampled by the meta-controller and trained on the given training data. The performance of the sampled network on the validation dataset is used as a reward score for updating the controller to sample better candidate networks from the search space in the next iteration.

A policy gradient method aims to approximate non-differentiable reward functions to train a model that needs parameter gradients (e.g. a network architecture). Zoph et al. \cite{zoph2016neural} used a recurrent neural network (RNN) policy controller trained with policy gradients to generate a sequence of actions to design a network architecture. The original algorithm in \cite{baker2016designing} is performed on macro search space, which designs the entire network architecture at once. Such a huge search space results in a prohibitive computational cost. For example, the algorithm \cite{baker2016designing} is performed on 800 graphics processing unit cards (GPUs) in 28 days (22400 GPU-days) on CIFAR10 dataset. To alleviate the computational burden, Zoph et al. \cite{zoph2018learning} proposed micro search space and adopted proximal policy optimization to optimize the RNN controller. Since the micro search space greatly reduces the size of architecture search space, the original algorithm in \cite{zoph2018learning} consumes 1800 GPU-days on CIFAR10 dataset. The Mnas algorithm \cite{tan2019mnasnet} proposed a factorized hierarchical search space that can not only enable layer diversity but also strike balance between the size of search space and flexibility. In addition, Mnas follows the same search strategy as in \cite{zoph2018learning} that automatically searches models to maximize the accuracy and minimize the real-world latency on mobile devices.

Q-learning \cite{watkins1989learning} is another class of popular RL-based NAS methods. The MetaQNN algorithm \cite{baker2016designing} adopted Q-learning to optimize a policy that can sequentially select a type of layer’s operation and hyperparameters of the network architecture. Zhong et al. \cite{zhong2018practical, zhong2020blockqnn} leveraged Q-learning with the epsilon-greedy strategy to search architectural building blocks. The building block is repeated several times and stacked sequentially to generate deeper network architectures for evaluation.

\subsection{EA-based NAS}

\begin{figure}[H]
\centering 
\includegraphics[width=0.45\textwidth]{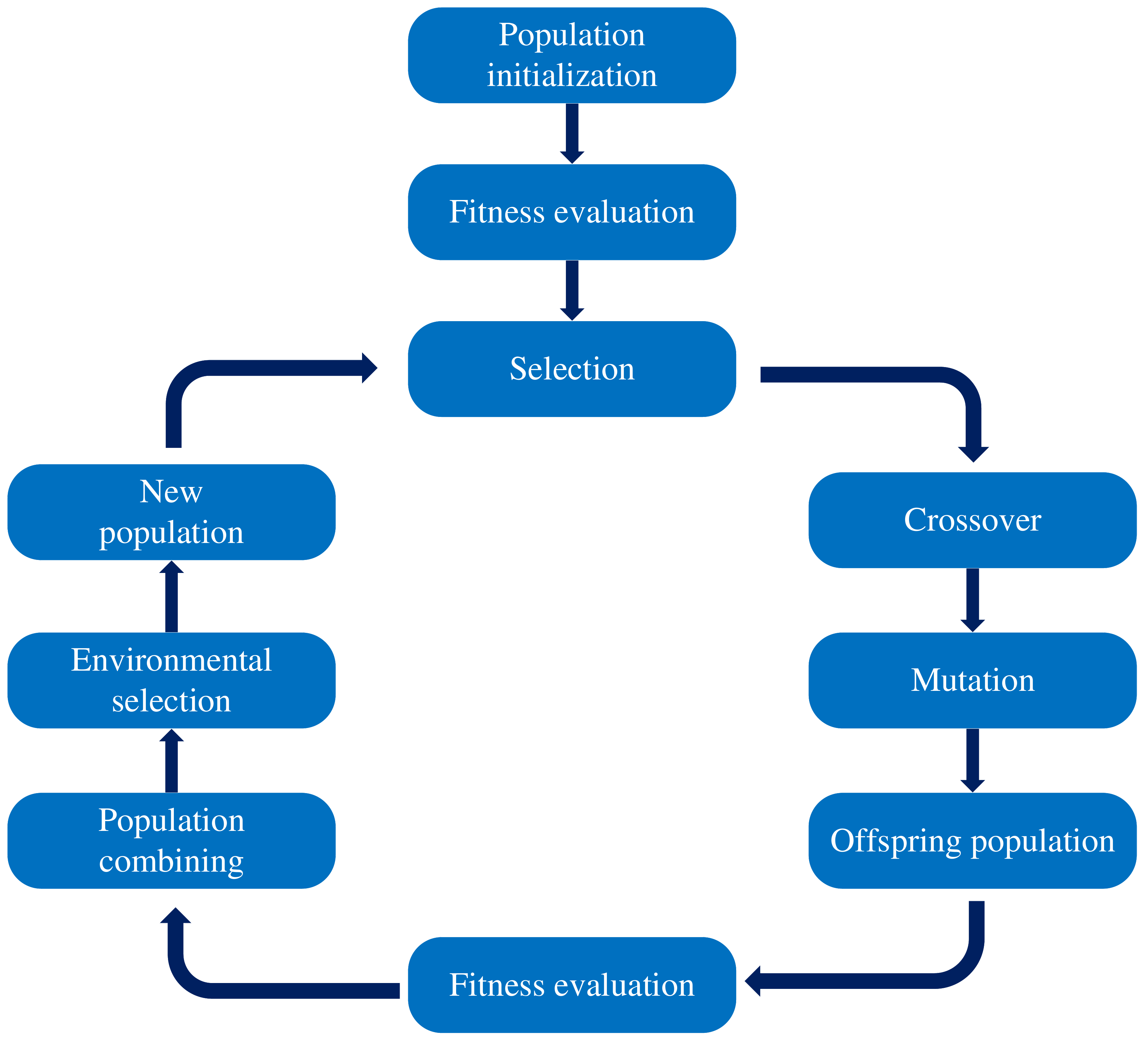} 
\caption{An overall framework of EA-based NAS algorithm.} 
\label{Fig_3}
\end{figure}

EA-based NAS is another stream of NAS methods. EA is a type of population-based heuristic computational paradigm \cite{schmitt2001theory}. The individual inside the population represents a candidate solution to the investigated problem. In EA-based NAS, the evolutionary algorithm is adopted as the search strategy to search for the best network architecture. Hence, each individual in the population should represent a candidate network architecture. With the evolution of the population, the performance of the network architecture is getting better and better on the given datasets. The overall framework of EA-based NAS algorithm is shown in Fig.\ref{Fig_3}.

The workflow of EA-based NAS follows several steps: firstly, a population requires to be randomly initialized with the pre-defined population size by the associated phenotype-to-genotype mapping strategy. Each individual is decoded to a neural network and iteratively trained on the given train dataset for several epochs. The fitness value of the individual is the accuracy of the validation dataset. All individuals will participate in the evolutionary process. Secondly, the selection strategy (e.g. tournament selection \cite{miller1995genetic}) is adopted to choose the parent individuals according to the fitness value, and the crossover and mutation operators are applied to the chosen parents to generate the new offspring until achieving the pre-defined size. Thirdly, the environmental selection is performed on the combined population to select the better individuals surviving into the next generation. This evolutionary process is repeated until the pre-defined termination condition is satisfied.

Traditional EA-based NAS methods, from the 1980s, used EAs to optimize the topology, weights, and hyperparameters of artificial neural networks, which is called neuroevolution \cite{schaffer1992combinations, yao1999evolving, stanley2002evolving, inden2012evolving}. Since the architecture of DNNs is more complex and has a large number of hyperparameters and connection weights, such traditional methods are not well suited for optimizing DNNs. As a result, recently, EA-based NAS methods only focus on optimizing the architecture of DNNs. The optimal weights for each candidate network are usually obtained by the gradient-based optimization algorithm \cite{bengio2012practical}.

Early work on optimizing CNNs, i.e. Genetic CNN \cite{xie2017genetic}, CoDeepNEAT \cite{miikkulainen2019evolving, liang2018evolutionary}, CGP-CNN \cite{suganuma2017genetic}, and CNF \cite{saxena2016convolutional}, have shown powerful performance. Researchers found that EAs can generate high-quality optimal solutions by using biologically inspired operators, including selection, crossover and mutation \cite{mitchell1998introduction}. Real et al. \cite{real2017large} adopted a variable-length encoding strategy to represent the architecture of CNNs and proposed a novel and intuitive mutation operators to explore the search space. EvoCNN algorithm \cite{sun2019evolving} used EAs to evolve the network architectures and corresponding initial connection weights at the same time. More effective initial connection weights can avoid neural networks falling into local optimal solutions. Sun et al. \cite{9075201, 8742788} proposed a variable-length encoding strategy to search the optimal depth of the CNN architecture through the basic genetic algorithm (GA). Damien et al. \cite{9439793} designed a new EA-based NAS framework, Matrix Evolution for High Dimensional Skip-connection Structures (ME-HDSS), to automatically remove the skip-connection structures in the DenseNet \cite{huang2017densely} model to further reduce the trainable weights and increase the performance of the model. Zhang et al. \cite{zhang2021adaptive} adopted I-Ching Divination Evolutionary Algorithm (IDEA) to optimize the complete network architecture, including the number of layers, the number of channels, and connections between different layers. In addition, to improve the search efficiency, the reinforced-based operator controller is developed to choose the different operators of IDEA. Liu et al. \cite{liu2020block} introduced multiple latency constraints in the architecture search process and then proposed latency EvoNAS (LEvoNAS) for optimizing network architecture. Sun et al. \cite{sun2018evolving} adopted EAs to design unsupervised DNNs for efficiently learning meaningful representations.

In fact, real-world tasks also need to consider multiple conflicting objectives, including floating-point operations per second (FLOPs) \cite{lu2019nsga, lu2020multi, lu2021neural}, latency \cite{guo2020single, lu2021neural}, memory consumption \cite{lu2021neural, elsken2018efficient}, inference time \cite{kim2017nemo}, among others. For example, the low power consumption and the high performance of the model in mobile applications. Compared with RL algorithm, EAs are more suited to solve multiobjective optimization problems in NAS. NEMO \cite{kim2017nemo} is one of the early studies using the elitist non-dominated sorting genetic algorithm (NSGA-II) \cite{deb2002fast} to minimize the inference time of a network and maximize the classification performance. Lu et al. \cite{lu2019nsga, lu2020multi} proposed NSGANetV1 method, which is formulated to design networks based on marco search space and micro search space with high classification performance and lower FLOPs using the NSGA-II algorithm. Wen et al. \cite{wen2021two} proposed a two-stage multi-objective EA-based NAS algorithm that can optimize the network architecture in transfer learning. In addition, multiobjective EA-based NAS also can be used in the federated learning community to reduce communication costs \cite{zhu2019multi, zhu2021real}. Note that federated neural architecture search is also a promising emerging research topic \cite{zhu2021federatednas}.

\section{Proxy methods in NAS}

The search procedure itself is laborious mainly because the training and evaluation candidate networks over a large search space are time-consuming. Therefore, more recently, many algorithms have been proposed for reducing computation costs and improving the efficiency of NAS with proxy methods such as low-fidelity estimation \cite{zhou2020econas, li2017hyperband, zhong2018practical, real2019aging, zhong2020blockqnn}, one-shot NAS \cite{zhang2020one, zhu2021real, zhang2022evolutionary}, and network morphism \cite{fang2020fna++, cai2019once}.

\subsection{Low-fidelity estimation}

Early methods in NAS have attempted to accelerate candidate neural network training and evaluation by low-fidelity estimation, such as adopting shorter training times (also denoted as early stopping strategy) \cite{zhong2020blockqnn, zela2018towards, baker2016designing, yang2022accelerating}, using lower-resolution of input images \cite{chrabaszcz2017downsampled}, starting with a small-scale dataset \cite{zoph2018learning}, using a subset of the full training set \cite{klein2017fast, 9354953, moser2022less}, and downscaling the size of candidate networks (e.g., reducing the channels of candidate neural network) \cite{liu2018darts, real2019aging, 9354953}. The low-fidelity estimation is adopted as surrogate measurements to guide the network architecture search process. Compared with full training optimization, low-fidelity estimation needs an order of magnitude fewer computation costs. NASNet \cite{zoph2018learning} designed small top-promising building blocks on a small-scale dataset to reduce the search cost. The optimized block module usually owns competitive generalization capabilities and can be transferred between different datasets or computer vision tasks. For example, the block networks designed on the CIFAR10 dataset can also achieve competitive performance in the ImageNet classification task \cite{deng2009imagenet} and MS-COCO object detection task \cite{lin2014microsoft}. Zhong et al. \cite{zhong2020blockqnn} used an early stopping strategy to enable the search strategy with fast convergence and reduce the search costs to 20 GPU-hours on the CIFAR10 dataset.

Although such low-fidelity estimation methods can save the search time, recent studies \cite{yang2020cars, zhou2020econas} indicated that they can lead to inaccurate evaluation of candidate networks, especially for complicated and large network architecture. For example, NASNet \cite{zoph2018learning} added an additional reranking stage before choosing the best network architecture, which trains the top 250 promising networks for 300 epochs each. The best network was ranked the 70th among 250 top promising neural networks according to performance ranking at low-fidelity estimation. Hence, simple low-fidelity estimation methods may result in low correlation in the prediction of the performance of the network. Zhou et al. \cite{zhou2020econas} also discussed this phenomenon and studied the impact of different low-fidelity estimation methods on the performance ranking of neural network architectures. Damien et al. \cite{o2021evolutionary} provided an analysis of the correlation between different lower fidelity estimation methods and final test performance.

\subsection{One-shot NAS}

\begin{figure}[H]
\centering 
\includegraphics[width=0.45\textwidth]{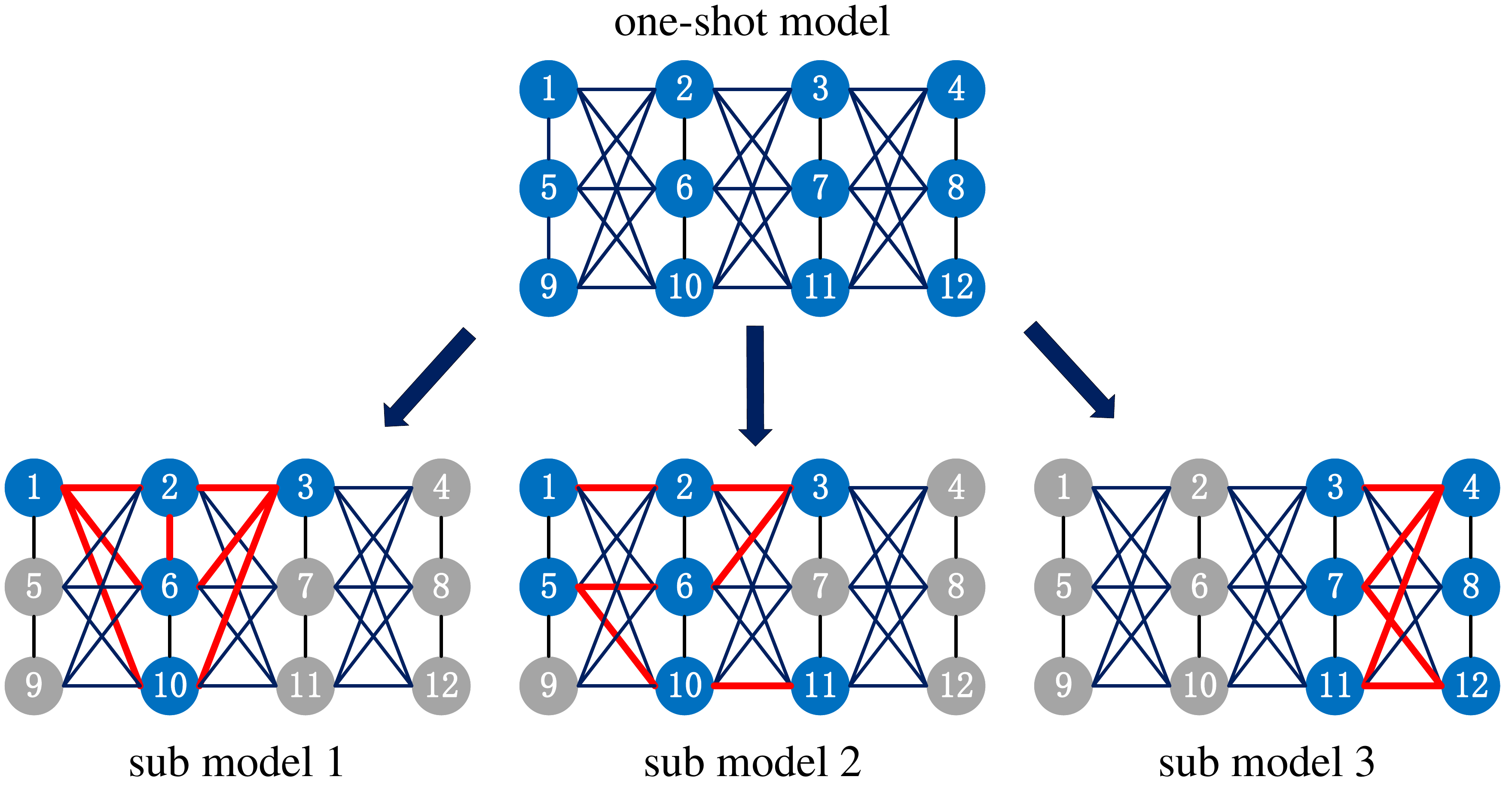} 
\caption{An example of the search space based on the one-shot model. Blue nodes represent active nodes. Grey nodes represent inactive nodes. The red lines denote the active paths in the one-shot model.} 
\label{Fig_4}
\end{figure}

Apart from assessing network candidates under low-fidelity estimations, one-shot NAS methods, also called weight sharing methodologies, have become more and more popular recently. One-shot NAS aims to construct a single large neural network (also denoted as the one-shot model) to emulate any network in the search space. As is shown in Fig.\ref{Fig_4}, the one-shot model can be viewed as a DAG, all possible architectures are different sub-graphs of the one-shot model. The node in the DAG represents a layer (e.g. convolution layer) in the neural network, and directed edges stand for the flow of information (e.g. feature maps) from one node to another. Once the one-shot model training is complete, all network candidates in the search space can directly inherit weights from the one-shot model for evaluating performance on the validation dataset rather than training thousands of separate sub-models from scratch. The workflow of one-shot NAS follows four sequential steps:

\begin{itemize}

\item Design a one-shot model as the search space that contains all possible network architectures. 

\item Train the one-shot model to convergence by a network sampling controller. 

\item Adopt a search strategy (e.g. an evolutionary algorithm, reinforcement learning, and the gradient-based method) to find the best sub-model based on the pre-trained one-shot model. 

\item Fully train the best submodel from scratch and evaluate its performance on the test dataset.

\end{itemize}

The source of “weight sharing” is first proposed from the ENAS \cite{pham2018efficient}, which used the RL-based method as the search strategy and forced all sub-models to share parameters from the one-shot model to avoid training each sub-model from scratch. Compared with NASNet \cite{zoph2018learning}, ENAS reduces the search cost from 1800 GPU-days down to 16 GPU-hours on CIFAR10 classicifation task. Luo et al. \cite{luo2018neural} proposed the NAO algorithm, which replaces the RL-based search strategy of ENAS \cite{pham2018efficient} with a GD-based auto-encoder that directly exploits weight sharing. Gabriel et al. \cite{bender2018understanding} analyzed the role of weight sharing in ENAS. Zhang et al. \cite{zhang2020efficient} proposed SI-EvoNAS algorithm, an EA-based one-shot NAS framework, which jointly optimizes the network architecture and associated weights. In SI-EvoNAS, a sampling training strategy is proposed to train the parent individuals. A node inheritance strategy is proposed to generate the offspring individuals, which can force the offspring individuals to inherit the weights from their parents, thereby avoiding training the offspring individual from scratch. 

\begin{figure}[H]
\centering 
\includegraphics[width=0.45\textwidth]{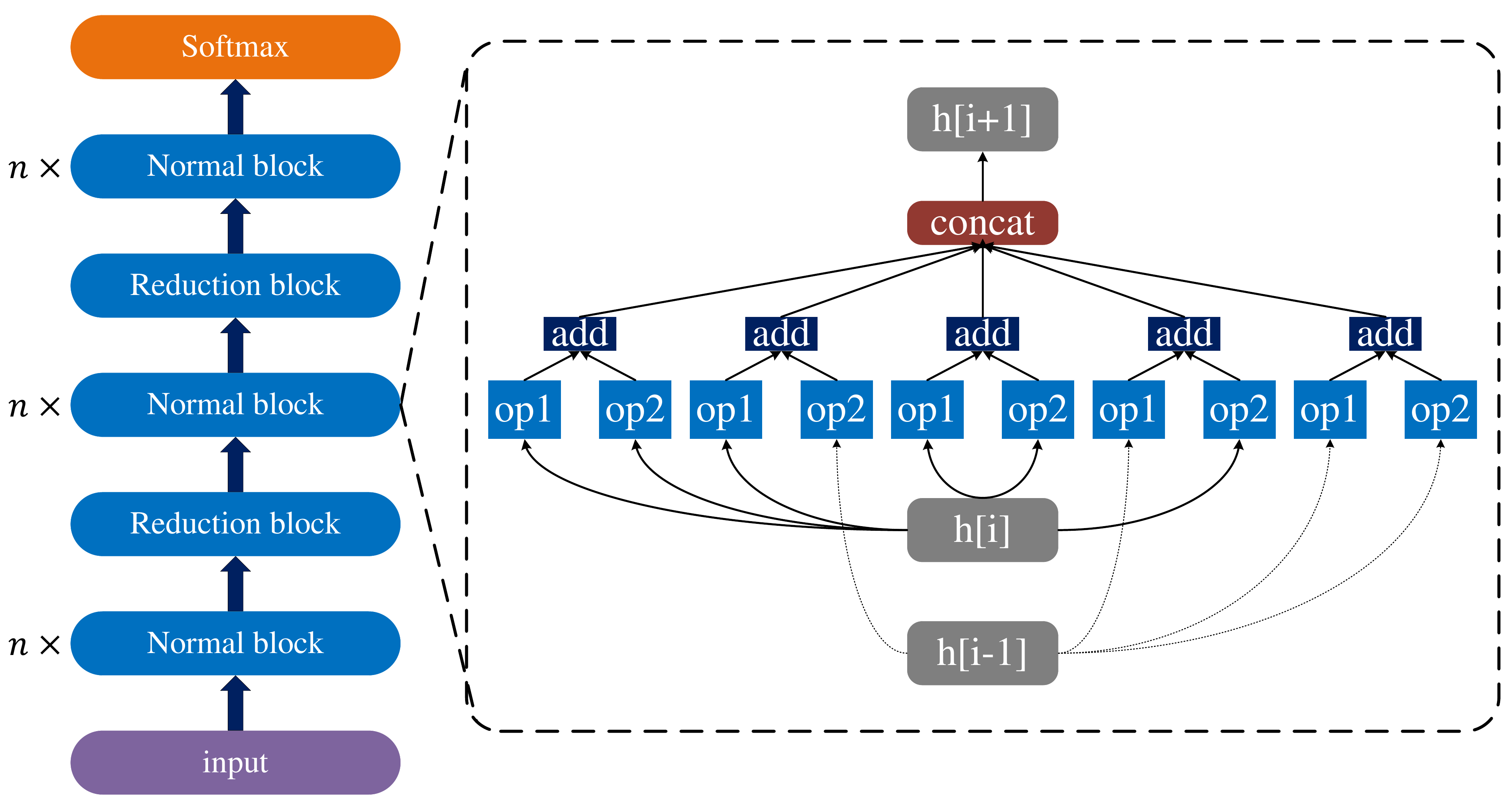} 
\caption{An exemplar relaxation trick based on micro search space.} 
\label{Fig_5}
\end{figure}

Yassine et al. \cite{benyahia2019overcoming} discovered the “multi-model forgetting” phenomenon in the one-shot NAS. Specifically, since the shared weights of the nodes are overwritten during the one-shot model training, the performance of the previously trained network will degrade when optimizing the network for subsequent training. As a result, the confidence of the ranking of candidate networks is no longer reliable. To solve this problem, Yassine et al. proposed a  statistically-justified weight plasticity loss to regularize the one-shot model training. Bender et al. \cite{bender2018understanding} proposed a “path dropout” algorithm that randomly removes some nodes of a one-shot model according to the dropout rate throughout one-shot model training. Guo et al. \cite{guo2020single} proposed a uniform sampling strategy to ensure all nodes in the search space have equal optimization opportunities during the one-shot model training. Chu et al. \cite{chu2021fairnas} proposed the FairNAS algorithm, which unveils the root cause of the effectiveness of the one-shot model training under two fairness constraints including strict fairness and expectation fairness. Zhang et al. \cite{zhang2020one} formulated the training process of the one-shot model as a constrained continual learning optimization problem to ensure the current network training does not degrade the performance of previous networks. To achieve this, Zhang et al. \cite{zhang2020one} proposed a search-based architecture selection (NSAS) loss function for one-shot model training. Ding et al. \cite{ding2021bnas} designed broad scalable network architecture to avoid performance drops in ENAS during the phase of network performance estimation. Ning et al. \cite{zhou2022close} introduced an operation-level encoding scheme to change the parameter sharing scheme dynamically.


Another family of one-shot NAS algorithms adopts a continuous relaxation trick to transfer the discrete search space to be a continuous space by approximating the connectivity between different layers in the networks through real-valued variables. And then, the real-valued variables and the model weights are jointly optimized by the gradient descent (GD) procedure at the same time. Fig.\ref{Fig_5} shows an exemplar relaxation trick based on micro search space. The first work of relaxation-based NAS methods is generally viewed as the Differentiable NAS (DARTS) \cite{liu2018darts}, which employed the GD method to search the best building blocks of a CNN architecture, and the experimental results on CIFAR10 and ImageNet have presented its performance. However, DARTS requires excessive GPU memory during architecture search because all layer operation candidates have to be explicitly instantiated in the memory \cite{9354953}. Hence, Chen et al. \cite{chen2019progressive} reduced the size of the one-shot model to solve this issue. Stochastic neural architecture search (SNAS) \cite{xie2018snas} adopted gradient information from generic differentiable loss to further accelerate the process of the network search. Cai et al. \cite{cai2018proxylessnas} proposed ProxylessNAS, which employed binary gates (1 or 0) to convert the architecture parameters of the one-shot model into binary representations. During the architecture search, ProxylessNAS only active a single path in the one-shot model by the binary gates. Therefore, the computational cost of ProxylessNAS is roughly the same as that of training a single network. Benefiting from the low computational cost, ProxylessNAS can directly search for an optimal network architecture on ImageNet. Zhou et al. \cite{zhou2021exploiting} adopted a high-order Markov chain-based method to combine current indicators of an operation and its historical importance, which can provide more accurate decisions of the importance of the operation.

Wang et al. \cite{9525822} observed that DARTS would suffer severe degradation because of the mechanisms in training and discretization. Hence, DARTS tends to keep more skip-connect operations in the final optimal model and converge to shallow network architecture. To address the issue aforementioned, Xu et al. \cite{xu2019pc, 9354953} proposed partially-connected DARTS (PC-DARTS) algorithm by using a channel sampling scheme that samples a subset of channels in each training step instead of activating all channels to reduce the redundant space of the one-shot model during the search process. iDARTS \cite{9525822} adopted the node normalization strategy to maintain the norm balance of different nodes, thereby avoiding the degradation in DARTS. Zhou et al. \cite{zhou2021bayesian} proposed a BDAS algorithm based on DARTS to search a domain matching diagnostic network. To solve the unfairness problem of one-shot model training in DARTS, the BDAS proposed a corresponding strategy based on path-dropout and warm-up and adopted variational Bayesian inference to estimate the uncertainty of model matching. 

Benefiting from the significant improvement in computational efficiency of one-shot NAS algorithms and significant progress in image classification tasks, the application of one-shot NAS in the task of object detection has also attracted the attention of researchers. Early methods, such as NASNet \cite{zoph2018learning} and RelativeNAS \cite{9488309}, transferred the searched network based on the classification tasks as the backbone for object detectors without further searches, which cannot guarantee optimal adaptation for any detection task. With the development of NAS in automating the design of network architecture, it has also boosted the research into automatically designing the network architecture for object detectors rather than handcraft design. Inspired by one-shot NAS methods, OPANAS \cite{liang2021opanas} proposed a novel search space of an FPN structure based on the one-shot model. Then, an EA-based one-shot NAS method is adopted to find the optimal path in the one-shot model to construct the FPN structure. Xiong et al. \cite{xiong2021mobiledets} proposed MobileDets for mobile or server accelerators. The authors wanted to replace depth-wise convolutions with regular convolutions in inverted bottlenecks through the one-shot NAS method to improve both the accuracy and latency of models. Hit-Detector \cite{guo2020hit} adopted the gradient-based NAS method \cite{liu2018darts} to optimize the architecture of all components of the detector. The estimation quality of various one-shot and zero-shot methods was systematically investigated on five NAS benchmarks \cite{ning2021evaluating}.

\subsection{Network morphism}
Parameter remapping strategy is a class of popular methods to improve the computational efficiency in NAS. The parameter remapping strategy, also called network transformation/morphism, aims to remap the parameters of one model to each new model to speed up the training process of the new model. Chen et al. \cite{chen2015net2net} tried to transfer the trained parameters from a small model to a new larger model with the help of the concept of function-preserving transformations, which effectively improves the performance of the large model on the ImageNet classification task. Following this manner, Cai et al. \cite{cai2017reinforcement} adopted the Net2Deeper and Net2Wider operators from \cite{chen2015net2net} during the architecture search. EAS \cite{cai2018efficient} adopted network morphisms to grow the width of the layer and the depth of the network. Cai et al. \cite{cai2019once} proposed Once-for-All method, which adopts a progressive shrinking algorithm to train the one-shot model. After the one-shot model has been trained, Once-for-All maps the parameters of the one-shot model to sub-models directly.

\section{Surrogate-assisted NAS}

While proxy methods try to make an estimation of the network performance according to a set of sub-optimal weight matrices, the surrogate-assisted NAS approaches take advantage of those trained networks by modeling a surrogate. Generally, the surrogate models are trained by a set of training data, which consists of network encoding and the associated performance in pairs. The trained surrogate can evaluate the performance of any candidate architecture in the search space, averting the huge computational overhead of training networks with poor performance. This section begins with a brief introduction to surrogate-assisted optimization before digging into different types of surrogate-assisted NAS techniques, such as Bayesian optimization-based NAS and surrogate-assisted evolutionary algorithm-based NAS. Then we give a detailed discussion about federated NAS and multi-objective optimization in surrogate-assisted NAS methods.

\subsection{Surrogate-assisted Optimization}

Surrogate-assisted evolutionary optimization derives from the practical challenges that some engineering optimization problems have no analytic objective functions, or the evaluation of a candidate solution can take hours or even a few days \cite{jin2011surrogate, jin2005comprehensive}. For example, in aerodynamic design optimization, the numerical simulations could be very time-consuming \cite{tao2019application}. When population-based optimization methods, such as EAs, are adopted to search for the optimal solution for a given task, a huge number of candidate solutions need to be evaluated. Consequently, the calculation cost under these circumstances is unaffordable. In order to cope with this, surrogate-assisted approaches investigate training surrogate models based on a limited amount of data, and provide reliable evaluations for candidate solutions within optimization. The limited training data often comes from physical experiments, numerical simulations, or historical information \cite{sun2018semi, wang2021transfer, wang2016data}.

The data for training surrogate models has two categories in general: direct data and indirect data \cite{jin2018data, sun2021data}. Direct data consists of at least two parts: decision variables and the corresponding objective or constraint values, which can be directly adopted to train a surrogate model. Most of the surrogate-assisted NAS approaches belong to this category, since the encoding of network architectures can be regarded as decision variables, and the network performance is taken as objective values. However, in other cases, it may be impossible to collect data in the form of decision variables and objective function values. For example, in trauma system design optimization, there is no mathematical formulation of the objective function, and the emergency accident records are the only information accessible \cite{wang2016data}. Under these circumstances, the designer should calculate the objective values from the indirect data before building and training a surrogate model.

Depending on whether new sample points can be collected by making evaluations of candidate solutions with a real objective function during the optimization process, surrogate-assisted evolutionary optimization can be categorized into offline and online algorithms. In offline surrogate-assisted evolutionary optimization, a set of candidate solutions together with their ground-truth objective values are collected to train a surrogate model before the optimization starts. During the optimization process, no new data can be actively generated for real evaluation, and only the predictions provided by the trained surrogate model can guide the evolutionary search. In other words, the final performance of the algorithm is mostly determined by the distribution of samples in the offline training set as well as the accuracy of the surrogate model. In practice, it is non-trivial to collect a set of ideal offline data in terms of quality, quantity and distribution. The data can even be noisy or incomplete in some circumstances. These challenges may hinder the offline algorithms from achieving better performance. To tackle this problem, some techniques can be adopted to improve the model performance of offline surrogates \cite{sun2021data}. From the perspective of data, we can use data preprocessing and data mining techniques to alleviate the influence of noise and uneven distribution on data quality. From the perspective of surrogate models, we can either choose an appropriate model elaborately for a specific task, or build an ensemble of surrogate models to improve its robustness. From the perspective of tasks, multi-fidelity fitness evaluation strategies and knowledge transformation between similar tasks can also be adopted to reduce the computational overhead.

By contrast, in online surrogate-assisted evolutionary optimization, there are no more restrictions on training surrogate models with limited offline data. During the evolutionary search, new data points can be actively sampled and evaluated by the objective function to get ground-truth labels. Consequently, we can enrich the quantity and quality of the training set of the surrogate model, thus improving its prediction accuracy. In addition to the techniques adopted in offline algorithms, we also need model management concerned with the sampling number, frequency and selection criteria to strike a balance between exploration and exploitation.

Model management is a key factor in online surrogate-assisted optimization, since it enables more efficient evaluations of the objective function as well as a more accurate surrogate approximation \cite{jin2018data}. In surrogate-assisted evolutionary optimization, fitness evaluations are provided by surrogate models instead of the real objective functions, contributing to a significant reduction in computational overhead. However, it is often infeasible to rely solely on surrogates for the fitness approximation. With a high-dimensional objective function and limited training data, a surrogate model without much a priori knowledge of the problem itself could have a bad performance. Model management, based on individuals or generations, investigates how to use the original fitness function efficiently in a surrogate-assisted optimization process. For individual-based strategies, the best individual or randomly selected individuals in each generation will be evaluated by the original function, while others will be evaluated by the surrogate model. For generation-based strategies, the real evaluation of all individuals in the population will be carried out every few generations. 

\subsection{Bayesian Optimization in NAS}

Bayesian optimization (BO) has emerged as a powerful tool for solving expensive black-box optimization problems, which has been widely applied to hyperparameter tuning of various machine learning scenarios such as recommendation systems, natural language processing, robotics and reinforcement learning \cite{shahriari2015taking, snoek2012practical}.

As a sequential strategy for derivative-free global optimization, the general principle of BO is to build a probabilistic model for the objective of interest, which can be updated using the collected data. During the optimization loop, the model provides an informative posterior distribution to guide the optimization directions and strike a balance between exploration and exploitation. 

The framework of BO has two key components: a probabilistic surrogate model and an associated acquisition function. The surrogate model contains our assumptions about the priori information of the unknown objective function. Gaussian Processes (GPs) are the most common choice for surrogate models. It is assumed that the distribution of the unknown objective function follows a Gaussian process, which means the function should be smooth and the deviations are Gaussian noises. The acquisition function is introduced to determine which is the next sample point to be evaluated. Since BO aims at finding the global optimum with fewer function evaluations, the optimization of the acquisition function is expected to compromise between exploration and exploitation. Specifically, those samples with a larger  degree of uncertainty (exploration) or a higher predicted value (exploitation) are preferred to be evaluated in sequence for a maximization problem. Different acquisition functions have been designed, such as probability of improvement, expected improvement \cite{mockus1978application}, the Gaussian process upper confidence bound \cite{srinivas2010gaussian} and entropy search \cite{hennig2012entropy}. It is worth noting that the acquisition function should be computationally much cheaper (but not necessarily easier) to optimize than the objective function, since the original black-box function is time-consuming and computationally intensive.

When searching for optimal network architectures from the Bayesian optimization perspective, we consider NAS as an expensive black-box optimization problem. Before the search process, a search space $\mathcal{A}$ with related to a specific task/dataset is defined in advance, containing all possible solutions $\alpha \in \mathcal{A}$ for the optimal architecture. Actually, it is non-trivial to fully explore the entire space $\mathcal{A}$ due to the insupportable overhead. An objective function $f(\alpha)$ indicates the performance metric of neural networks, e.g., the validation accuracy of a given architecture. It may take several hours to fully train a network for evaluating $f(\alpha)$. A lot of work has been dedicated to Bayesian optimization-based NAS approaches recently. 

Auto-Keras \cite{jin2019auto} is an open-source AutoML system. It's developed from a NAS framework based on Bayesian Optimization, which is designed to operate on a network morphism. In this approach, a Gaussian Process model is built and trained with the existing architectures and their performance. To cope with the challenge that the original network architectures are not in Euclidean space, a neural network kernel is proposed based on the edit-distance for morphing one architecture to another. The optimization of the acquisition function is also re-designed for the tree-structured search space in network morphism.
NASBOT \cite{kandasamy2018neural} is a Gaussian process based BO framework for neural architecture search. A distance metric named OTMANN (Optimal Transport Metrics for Architectures of Neural Networks) is developed to quantify the similarity between two networks in the search space, and the acquisition function is optimized by EA approaches. A main challenge for BO in a graph-like search space is how to capture the topological structures of neural networks. NAS-BOWL \cite{ru2020interpretable} combines Weisfeiler-Lehman graph kernel with a Gaussian process, enabling the surrogate model to be directly defined in a graph-based search space.
However, the above NAS methods are full-fledged, and we cannot tell which component makes the most contribution. White et al. \cite{white2021bananas} made a systematic analysis of the ``BO + performance predictor" framework with five separate components: the network encoding, the performance predictor, the uncertainty calibration, the acquisition function and the acquisition function optimization. Then a BO-based algorithm named BANANAS was proposed based on the analysis. Each network architecture in the search space is represented as a labeled DAG, and path encoding is developed to improve the predictor performance. 

While the BO-based NAS algorithms in \cite{jin2019auto, kandasamy2018neural, ru2020interpretable} focus on designing neural network kernels for Gaussian Process, other works introduced different kind of surrogate models. BONAS \cite{shi2020bridging} combines Graph Convolutional Network (GCN) \cite{kipf2016semi} and Bayesian sigmoid regressor as the surrogate model instead of GP. The individuals with the top-$k$ UCB scores will be selected and evaluated by weight-sharing in order to update the surrogate model. \cite{ma2019deep} developed a graph Bayesian optimization framework with a Bayesian graph neural network as surrogate model.
As the first Bayesian approach for one-shot NAS, BayesNAS \cite{zhou2019bayesnas} uses a hierarchical automated relevance determination prior to model architecture parameters, alleviating the inappropriate operation over $zero$ operations in most one-shot methods. 
One of the challenges for Bayesian Optimization in NAS is the high-dimensional and discrete decision space. To counter this problem, Neural Architecture Generator Operation (NAGO) \cite{ru2020neural} considered NAS as a search for the best network generator, and built a novel graph-based hierarchical search space which can cover a wide range of network architectures with only a few hyperparameters. Consequently, the problem dimensionality was greatly reduced, allowing Bayesian Optimisation to be used effectively. GP-NAS \cite{li2020gp} investigated the correlation among different architectures as well as the performance from a Bayesian perspective,where a kernel function for NAS is specially designed by categorizing operations into multiple groups.

In addition to direct search of network architectures, BO can also be integrated with other proxy approaches to improve search efficiency, such as knowledge distillation \cite{trofimov2020multi} and surrogate models \cite{wei2020npenas, cho2022b2ea}.

\subsection{Surrogate-assisted Evolutionary NAS}

As a class of population-based optimization algorithm, EAs maintain a population consisting of feasible solutions, and generate offspring with progressively better performance, enabling the algorithm to converge towards the optimal solution. In EA-based NAS approaches, individuals in the population are considered as candidate architectures defined in the search space, and the genotypes are determined by the network encoding \cite{white2020study}. The fitness function reflects the network performance, which can be the validation accuracy or other evaluation factors. Due to the superior performance of EAs in solving black-box optimization problems, much work has focused on designing EA-based NAS approaches.

In 2017, Google proposed the LargeEvo algorithm \cite{real2017large}, which uses a genetic algorithm to search for well-performed CNN architectures on CIFAR-10 and CIFAR-100 datasets. This is commonly thought to be the first evolutionary-based NAS algorithm. Since then, a lot of work has been dedicated to searching for optimal network architectures by EA approaches \cite{real2019regularized, liu2017hierarchical, xie2017genetic, sun2019evolving, zhu2021toward, lu2020multi, elsken2018efficient}.

However, it is non-trivial to directly use evolutionary algorithms to search for the optimal network architectures for a given task. 
First of all, the fitness evaluation of one candidate solution takes a considerable amount of training time. In order to get the performance evaluation of a candidate architecture, one should initialize the weights of the given network, and then train it on the training dataset by using the gradient descent approach over a large number of epochs before convergence. As the training set and network size grow, this procedure might take hours or even days. On the other hand, the EA approach always needs to evaluate a large number of candidate solutions at a time due to its population-based properties. As the task difficulty and the population size increase, this time-consuming process will be a challenge to limited computing resources \cite{liu2021survey}.

\begin{figure}[H]
\centering 
\includegraphics[width=0.45\textwidth]{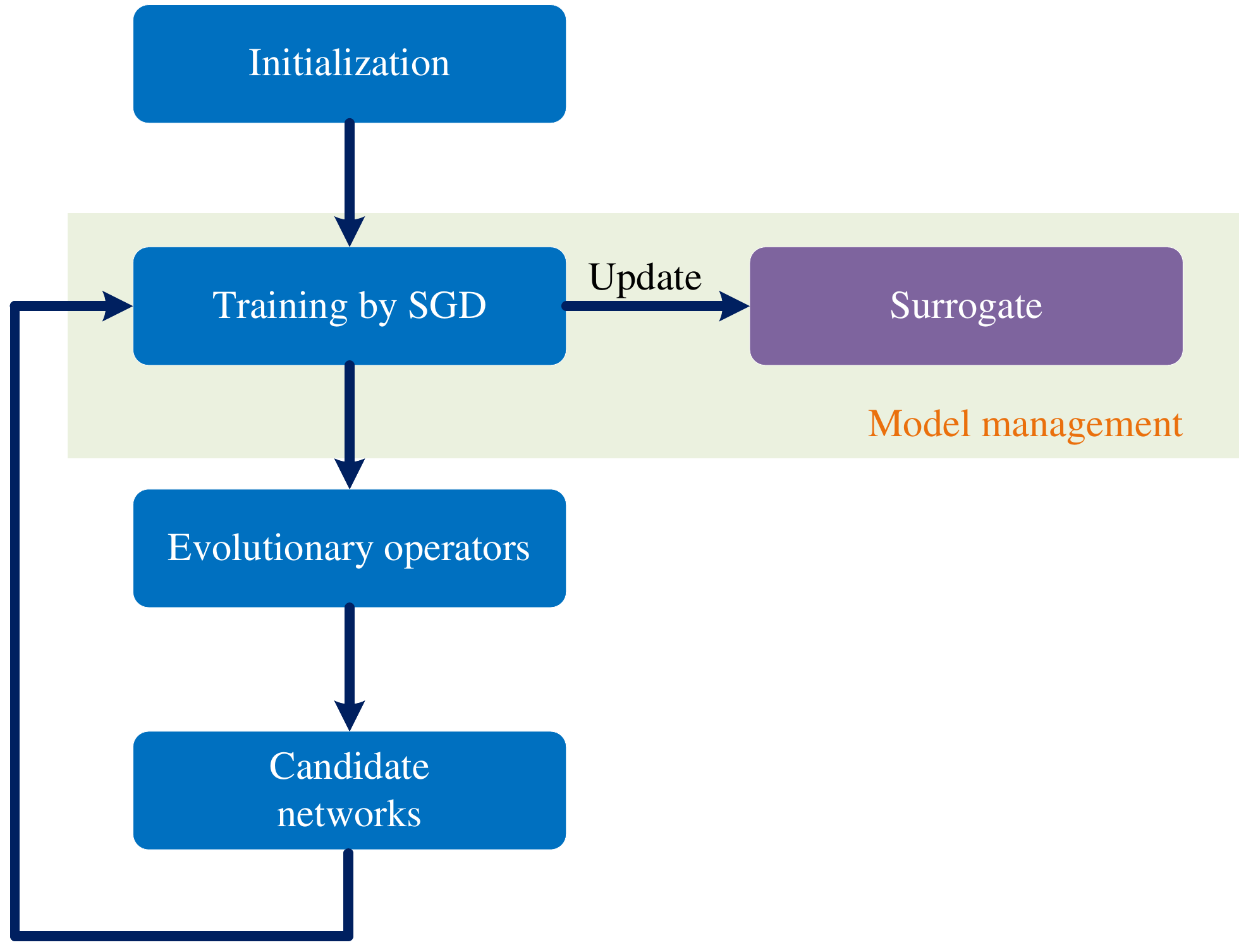} 
\caption{A basic framework of surrogate-assisted evolutionary NAS algorithms} 
\label{Fig_6}
\end{figure}

To address this problem, a performance predictor (also known as a surrogate model) is introduced \cite{jin2011surrogate}. The surrogate model aims to estimate network performance without training the network from scratch. The basic steps of surrogate-assisted evolutionary NAS approaches are:

\begin{itemize}

\item Sample a set of network architectures from the search space, and train them from scratch to get the ground truth labels. Store the samples in an archive $A$.

\item Use the archive $A$ to construct a training dataset $D_{tr}$. Build and train a surrogate model $M$ by using $D_{tr}$.

\item Perform neural architecture search by EAs, with the network performance predicted by the surrogate model $M$.

\item Select candidate architectures for real evaluation according to the model management strategy. Update the archive $A$ and the training dataset $D_{tr}$.

\item Train and evaluated the selected architectures. Update the archive $A$ and the training dataset $D_{tr}$.

\item Update the surrogate model $M$ for the next iteration.

\end{itemize}

According to \cite{sun2019surrogate} and \cite{sun2021novel}, there are mainly four categories of network performance predictors: learning curve-based predictors \cite{domhan2015speeding, klein2016learning, baker2017accelerating, rawal2018nodes}, weight sharing-based predictors \cite{pham2018efficient}, shallow training-based predictors \cite{sun2019evolving} and end-to-end performance predictors \cite{sun2019surrogate, deng2017peephole, greenwood2022surrogate}. 
The learning curve-based approaches require a partial training of the neural network, and then extrapolate the upcoming trends of the curve based on the observed part. An example for learning curve-based models is presented in \cite{rawal2018nodes}, where LSTM is adopted as a seq2seq model to predict the network performance based on the learning curve of the first few epochs. 
Domhan et al. \cite{domhan2015speeding} built a set of parametric functions as a probabilistic model, and used it to extrapolate from the initial part of the learning curve to a future point. A run will be terminated if its performance prediction on validation set seems unlikely to achieve the best model performance so far. Based on this work, Klein et al. \cite{klein2016learning} developed a Bayesian neural network as the probabilistic model, and improved its prediction by a specially-designed learning curve layer. Baker et al. \cite{baker2017accelerating} trained sequential regression models instead of a single Bayesian model to estimate the validation accuracy of candidate networks. However, the learning curve-based methods still rely on the training process of the candidate networks, making the prediction time-consuming. Similarly, the shallow training-based and weight sharing-based predictors (as mentioned in Section III-B) may also suffer from poor generalization ability due to insufficient training and weight dependency assumption.

Another option is to build and train an end-to-end predictor, which directly takes a network architecture as input, and predicts the network performance as its output.  Deng et al. developed an end-to-end training approach called \textit{Peephole} \cite{deng2017peephole}. It uses LSTM to encode the individual layers as well as the number of epochs into a structural embedding, and feds it into a multi-layer perceptron to predict the validation accuracy after the given epoch. In contrast to the learning curve-based methods, \textit{Peephole} can predict the entire learning curve without knowing the initial part of it. Although this end-to-end training can make the prediction more efficient, training such a surrogate model requires a large number of pre-trained neural networks, which introduces additional computing overhead.

To further reduce the computational overhead, E2EPP \cite{sun2019surrogate} is one of the representative end-to-end performance predictor models in surrogate-assisted NAS, which can achieve good performance with limited training data. In E2EPP, a random forest \cite{ho1995random} is adopted as a surrogate model for predicting the network performance by given the network encoding. The proposed performance predictor is then embedded with AE-CNN \cite{sun2019completely}, an evolutionary deep learning method to further verify its effectiveness. The advantage of adopting a random forest as a proxy model instead of other models like neural networks is that it can directly accept discrete data as input, without the need for a large amount of training data. As an ensemble learning method, a random forest model consists of a set of decision trees. Each decision tree will select a subset of features from the decision variables randomly. During the training process, each decision tree learns a mapping from the features to the targets. During the prediction process, the outputs of selected decision trees are averaged as the final prediction of the random forest.
NPENAS \cite{wei2020npenas} is another neural predictor guided approach for evolutionary neural architecture search. In NPENAS, multiple offspring are generated from each parent instead of one, in order to enhance the exploration ability of the search algorithm. To alleviate the additional computational cost caused by extra candidate architectures, a neural predictor is introduced to rank the offspring from the same parent, and only the candidate with the highest predicted performance will be selected for real evaluation. Two different kinds of neural predictors are proposed in NPENAS. The first one is an acquisition function defined by a graph-based uncertainty estimation network. Inspired by BO-based NAS algorithms, the authors assume that the distribution of network architectures in the search space is independent and identical, thus the performance prediction function can be regarded as a Gaussian Process, which is defined by its mean and standard deviation functions. A graph-based uncertainty estimation network is trained as a surrogate model to provide the mean and standard deviation values for a given architecture. Specifically, the model uses GINs \cite{xu2018powerful} and MLPs for architecture embedding, followed by fully-connected layers to predict the mean and standard deviation values. The second neural predictor has a similar structure, while it predicts the architecture performance directly. The algorithms embedded with these two neural predictor are named NPENAS-BO and NPENAS-NP respectively, and they both showed promising performance on NASBench datasets \cite{ying2019bench, dong2020bench}.

Wen et al. \cite{wen2020neural} proposed a simple yet effective method with only three steps for surrogate-assisted NAS. It first trains a set of randomly sampled network architectures to get the validation accuracies. Then the set of architectures are used to train a regression model (which is GCN in this paper). Finally, the trained model gives prediction on a large number of random architectures, and only the top-$K$ candidates will be trained from scratch to find the optimal architecture. \textcolor{black}{A recent work proposed by Greenwood et al. \cite{greenwood2022surrogate} augments the original DeepNEAT \cite{miikkulainen2019evolving} by introducing a surrogate model and two-phase active learning paradigm. During the initialization phase, networks are trained and evaluated by the standard DeepNEAT. During the active learning phase, the surrogate model is used to make predictions for the population rather than training and evaluating them directly. PRE-NAS \cite{peng2022pre} proposes a representative selection scheme which enables it to train a well-performed performance predictor within an extremely limited number of training samples. B2EA \cite{cho2022b2ea} introduces two BO as surrogates for an EA-based NAS approach. The first BO controls the search space attentively, and the second BO predicts performance for network architectures without training process. MORAS-SH \cite{liu2022bi} introduces an online surrogate model to predict the high-fidelity performance of architectures as a helper-objective for adversarial robustness search. DAU-NAS \cite{ying2022multi} adopts a random forest as a surrogate model to directly predict the performance of each subnet in addition to the weight sharing strategy.}

Graph Neural Network (GNN) is another type of effective surrogate model in evolutionary NAS \cite{wu2020comprehensive}. GNN takes a graph as its input and updates the graph attributes using message-passing techniques such as graph convolution layers. The output of a GNN is a graph with the same connectivity, while the graph attributes are updated and each node embedding has already incorporated the information from its neighborhood. Because the structures of most neural networks can be naturally encoded as DAGs \cite{wen2020neural}, an increasing amount of work is dedicated to using GNNs as surrogate models in NAS.
Lukasik et al. \cite{lukasik2020neural} used a GNN as a graph encoder to map networks from a discrete graph space to a continuous vector space. The model is first trained on CIFAR-10 dataset in a supervised learning method, and then its prediction ability is evaluated on zero shot scenarios with unseen architectures. Different from the traditional fully supervised way of training surrogates, Tang et al. \cite{tang2020semi} proposed a semi-supervised performance predictor based on GNN. Firstly, both labeled and unlabeled data are fed into an auto-encoder to get the meaningful embedding. Then a relation graph is constructed based on the embedding, revealing intrinsic connections among similar architectures. Finally, taking both the embedding and the relation graph as the inputs, a GCN is trained to predict the network performance of these unlabeled samples. Ning et al. \cite{ning2020generic} proposed a general graph-based architecture encoder called GATES. It regards neural networks as data processing graphs, and different operations are modeled as information transformations. By encoding cell architectures into embedding vectors, GATES can improve the performance predictors for surrogate-assisted NAS methods in various search spaces. Kyriakides et al. \cite{kyriakides2022evolving} proposed an evolutionary-based method to search for GCNs as performance predictors to evaluate the relative ranking of various network architectures.


While most of the surrogate-assisted NAS approaches share similar protocols of training a surrogate model by mean square error criterion (MSE), Sun et al. \cite{sun2021novel} proposed a pairwise ranking indicator (PRI) for surrogate training in NAS. In this protocol, a PRI is adopted to train the surrogate model instead of the traditional MSE function. Concretely, given any two sample architectures, the ranking information between the two samples will be used to train a regression model. The trained model can be easily integrated into evolutionary NAS algorithms, since its predictions reflect the real ranking among candidate architectures and are consistent with EAs' selection criterion. Another similar work is ReNAS \cite{xu2021renas}, where the authors defined a pairwise ranking-based loss function for training the surrogate model instead of the traditional element-wise loss functions such as MSE. The implicit idea is that predicting the relative ranking between two architectures is more important in evolutionary search than predicting the true performance values. \textcolor{black}{Similarly, HW-PR-NAS \cite{benmeziane2022pareto} introduces a novel loss function to rank the architectures based on their dominant relations, which avoids using multiple surrogates to estimate different objectives. Arch-Graph \cite{huang2022arch} formulates NAS as an architecture relation graph prediction problem, and trains a pairwise relation predictor to give architecture relation on any given task embeddings.}

In addition to a single type of predictor model, White et al. \cite{white2021powerful} demonstrated that the surrogate performance can be significantly improved by combining different categories of neural predictors.

\subsection{Federated NAS}
Federated Learning (FL) \cite{mcmahan2017communication} is an emerging technique in machine learning, where a global model is trained using distributed datasets stored locally on various clients, without the requirement to transform the privacy data to a centralized server or a third party \cite{yang2019federated, zhu2021federated}. With the surge of interest in privacy preserving, there is an increasing demand for search for specific network architectures under the federated learning framework \cite{zhu2021from, zhu2021real, he2020towards, liang2021self, singh2020differentially, zhu2019multi, xu2020federated}. The challenges of federated NAS compared to centralized NAS may come from huge communication cost, unbalanced data distribution as well as real-time deployment.

In terms of whether the architecture search and deployment processes are coupled, federated NAS can be categorized into offline and online algorithms \cite{zhu2021from}. There are two separate parts in offline federated NAS: the first stage is searching for an optimal network architecture, and the second stage is training and deploying this searched network to all clients. One example for offline federated NAS is \cite{zhu2019multi}, where the NSGA-II \cite{deb2002fast} is adopted to search for a set of Pareto optimal solutions for a given task. Offline NAS indicates that only the final optimized architecture will be employed, and the performance of the candidate solutions generated during the search process do not matter much. However, in some practical scenarios of federated learning, the optimization and deployment of neural networks should be performed simultaneously. In other words, online federated NAS requires that not only the final optimized model should work well, but also the architectures generated during the search process should have acceptable performances. Zhu et al. \cite{zhu2021real} proposed a real-time federated evolutionary NAS algorithm, where a double-sampling technique is adopted to reduce the huge computational and communication cost in online federated NAS. Specifically, only a randomly selected subset of clients participate in the training of a candidate network, and only sub-models of the supernet are sampled and trained and in the search process. Liu et al. \cite{liu2022federated} developed a multi-objective convolutional interval type-2 fuzzy model for federated NAS to ensure medical data security.

\begin{figure}[H]
\centering 
\includegraphics[width=0.45\textwidth]{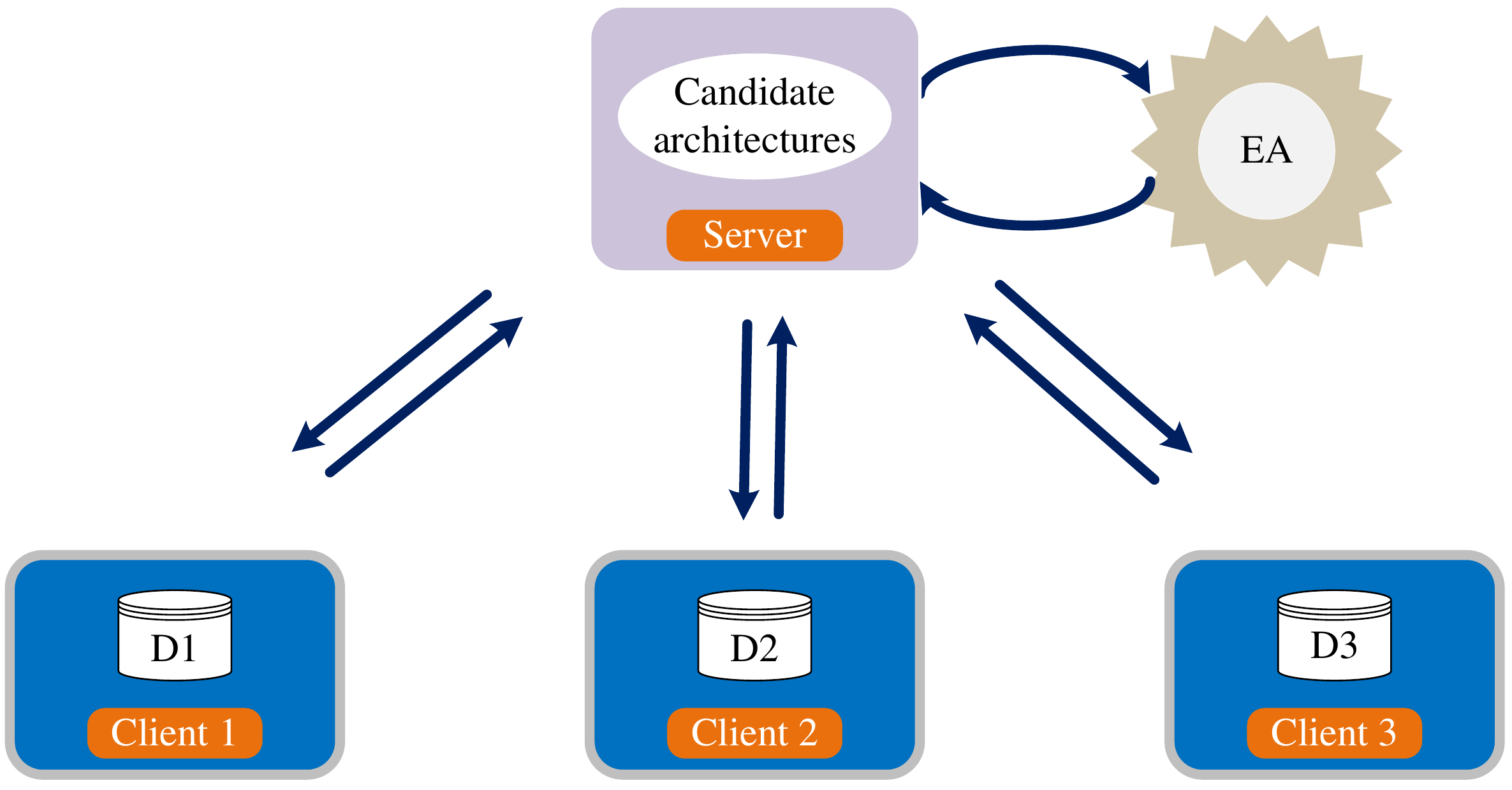} 
\caption{A basic framework for federated evolutionary NAS algorithms} 
\label{Fig_7}
\end{figure}

Another consideration for federated NAS is non-IID (non-independently and identically distributed) data on different clients \cite{zhu2021federated}. Due to the weight divergence of local models \cite{zhao2018federated}, non-IID data can lead to significant performance degradation when training a neural network under federated learning settings. To counter this problem, He et al. \cite{he2020towards} proposed FedNAS, a distributed NAS algorithm to search for optimal architectures in federated learning with non-IID data distribution. 

In federated NAS approaches, multiple clients collaboratively search for a well-performed network model without uploading their private data to a server. By keeping the data local, the security of client privacy is guaranteed to a certain extent. However, the gradient information of model parameters, which is exchanged during the training process, can still reveal privacy implicitly \cite{zhu2019deep}. In other words, avoiding data disclosure in federated NAS may be not enough in terms of privacy protection. To fill this gap, Singh et al. \cite{singh2020differentially} proposed DP-FNAS, which uses differential privacy \cite{dwork2014algorithmic} to further improve security by adding Gaussian noise to the gradient values before being sent to the server for aggregation.

Most of the existing federated NAS algorithms focus on searching network architectures in horizontal federated learning scenarios, where each participant has access to the same features and labels with high quality. This kind of scenarios are defined as horizontal federated learning (HFL). Contrary to this, vertical federated learning (VFL) means all the participants share the same sample ID space, but with different feature subsets. In most of the VFL cases, only one participant holds the label set. In order to collaboratively search for an optimal neural network with feature-partitioned local datasets, Liang et al. \cite{liang2021self} proposed a self-supervised federated neural architecture search (SS-VFNAS) algorithm under cross-silo settings. First, each client uses a self-supervised NAS approach to find a local optimal network with its own data, and then all clients perform a supervised NAS to enhance the local optimal model in the VFL framework.

Federated neural architecture search is an emerging topic in automated machine (AutoML) learning research, and there are still some open issues to be investigated, such as introducing surrogate models to improve search efficiency, and designing robust NAS approaches to defend adversarial attacks.

\subsection{Multi-objective NAS}

In most practical scenarios of NAS, the model performance (validation accuracy) is not the only thing that matters. Several hardware constraints also need to be considered when deploying a deep learning model, such as computing power, memory usage, inference time and communication cost. The generic aim of multi-objective neural architecture search is to find a well-performed network architecture with minimal model size, computational complexity or inference time.

However, it's non-trivial to find such an acceptable network architecture, since the multiple objectives to be optimized simultaneously are always in conflict with each other. One method is to convert the multi-objective optimization into a single objective by introducing weight hyperparameters. Another idea is to search for a set of feasible solutions, where various architectures maintain a trade-off among conflicting objectives. Consequently, population-based evolutionary algorithms seem to be a natural choice for multi-objective NAS, and surrogate models are also adopted to improve the search efficiency \cite{jin2008pareto, dong2018dpp, lu2020multi, lu2020nsganetv2, lu2022surrogate}. Dpp-net \cite{dong2018dpp} trains a surrogate function to predict the classification accuracy for different architectures in a population, and only the Pareto-based top-$k$ candidates are trained and evaluated. Instead of training a united surrogate model for all architectures, NSGANetV1 \cite{lu2020multi} down-scales the candidate networks and trains these proxy models to get the validation accuracy and FLOPs for non-dominated ranking. Following this work, NSGANetV2 \cite{lu2020nsganetv2} treats NAS as a bi-level optimization task and uses surrogates at both upper and lower levels. For the upper level (architecture) optimization, four different kinds of surrogate models are trained at each iteration, and then Adaptive Switching is used to choose the best model for performance prediction. For the lower level (weights) optimization, a supernet is trained once before the search process, and candidate architectures inherit weights from the supernet as a warm-start for the real evaluation.

\textcolor{black}{To conveniently investigate the comparisons, Table \ref{tab:accuracy} compares the performance of various computationally efficient NAS methods in terms of the classification accuracy and consumed GPU days on three popular datasets for image classification, namely CIFAR-10, CIFAR-100 and ImageNet.}

\begin{table*}[]
\centering
\caption{\textcolor{black}{Classification accuracy of various computationally efficient NAS approaches on CIFAR-10, CIFAR-100 and ImageNet}}
\label{tab:accuracy}
\begin{tabular}{ccccccc}
\hline
\multirow{2}{*}{Algorithm} &
  \multicolumn{3}{c}{Accuracy (\%)} &
  \multirow{2}{*}{GPU-days} &
  \multirow{2}{*}{Search method} &
  \multirow{2}{*}{\begin{tabular}[c]{@{}c@{}}Proxy method\\ / Surrogate\end{tabular}} \\ \cline{2-4}
 &
  CIFAR-10 &
  CIFAR-100 &
  ImageNet &
   &
   &
   \\ \hline
NAS v3 \cite{zoph2016neural} &
  95.53 &
  -- &
  -- &
  22400 &
  RL &
  -- \\
Genetic-CNN \cite{xie2017genetic} &
  92.9 &
  -- &
  72.13 &
  17 &
  EA &
  -- \\
AE-CNN \cite{9075201} &
  95.3 &
  77.6 &
  -- &
  27 &
  EA &
  -- \\
CNN-GA \cite{8742788} &
  96.78 &
  79.47 &
  -- &
  35 &
  EA &
  -- \\
Hier-EA \cite{liu2018hierarchical} &
  96.37 &
  -- &
  -- &
  300 &
  EA &
  -- \\
large-scale Evo \cite{real2017large} &
  94.60 &
  77.00 &
  -- &
  2750 &
  EA &
  -- \\
LEMONADE \cite{elsken2018efficient} &
  97.42 &
  -- &
  -- &
  90 &
  EA &
  -- \\
CGP-CNN \cite{suganuma2017genetic} &
  97.25 &
  -- &
  -- &
  227 &
  EA &
  -- \\ \hline
Amoebanet-A \cite{real2019regularized} &
  96.66 &
  81.07 &
  -- &
  3150 &
  EA &
  Low-fidelity estimation \\
MFENAS \cite{chen2022mfenas} &
  97.61 &
  -- &
  73.94 &
  0.6 &
  EA &
  Low-fidelity estimation \\
Block-QNN-S \cite{zhong2018practical} &
  96.70 &
  82.95 &
  -- &
  90 &
  RL &
  Low-fidelity estimation \\
MetaQNN (top model) \cite{baker2016designing} &
  93.08 &
  72.86 &
  -- &
  90 &
  RL &
  Low-fidelity estimation \\
PNAS \cite{liu2018progressive} &
  96.37 &
  80.47 &
  74.2 &
  225 &
  SMBO &
  Low-fidelity estimation \\
NASNet \cite{zoph2018learning} &
  97.35 &
  82.19 &
  74.0 &
  1800 &
  RL &
  Low-fidelity estimation \\
\begin{tabular}[c]{@{}c@{}}ME-HDSS \cite{9439793}\end{tabular} &
  93.65 &
  72.89 &
  -- &
  -- &
  EA &
  Low-fidelity estimation \\
\begin{tabular}[c]{@{}c@{}}EoiNAS \cite{9432795}\end{tabular} &
  97.5 &
  -- &
  -- &
  0.6 &
  GD &
  Low-fidelity estimation/ One-shot \\
\begin{tabular}[c]{@{}c@{}}ModuleNet\cite{9447766}\end{tabular} &
  97.33 &
  82.01 &
  78.69 &
  -- &
  EA &
  Low-fidelity estimation \\ \hline
ENAS \cite{pham2018efficient} &
  97.06 &
  -- &
  -- &
  0.5 &
  RL &
  One-shot \\
SI-EvoNAS \cite{zhang2020efficient} &
  97.31 &
  84.30 &
  75.8 &
  0.458 &
  EA &
  One-shot \\
Evo-OSNAS \cite{zhang2022evolutionary} &
  97.44 &
  84.16 &
  77.48 &
  0.5 &
  EA &
  One-shot \\
DARTS \cite{liu2018darts} &
  97.18 &
  82.46 &
  73.3 &
  1 &
  GD &
  One-shot \\
SNAS \cite{xie2018snas} &
  97.15 &
  79.91 &
  72.7 &
  1.5 &
  GD &
  One-shot \\
Proxyless NAS \cite{cai2018proxylessnas} &
  97.92 &
  -- &
  75.1 &
  -- &
  GD &
  One-shot \\
WPL \cite{benyahia2019overcoming} &
  96.19 &
  -- &
  -- &
  -- &
  RL &
  One-shot \\
BNAS \cite{9392299} &
  97.03 &
  -- &
  74.3 &
  0.19 &
  RL &
  One-shot \\
PDARTS \cite{chen2019progressive} &
  97.50 &
  84.08 &
  75.6 &
  0.3 &
  GD &
  One-shot \\
PC-DARTS \cite{9354953} &
  97.43 &
  82.89 &
  74.9 &
  0.3 &
  GD &
  One-shot \\
BayesNAS \cite{zhou2021bayesian} &
  96.98 &
  -- &
  73.5 &
  0.2 &
  GD &
  One-shot \\
GDAS \cite{dong2020searching} &
  97.25 &
  81.98 &
  74.1 &
  0.4 &
  GD &
  One-shot \\
\begin{tabular}[c]{@{}c@{}}RandomNAS-NSAS \cite{9247292}\end{tabular} &
  97.41 &
  82.44 &
  74.5 &
  0.7 &
  GD &
  One-shot \\
NAO+WS \cite{guo2021towards} &
  96.47 &
  -- &
  74.3 &
  0.3 &
  GD &
  One-shot \\
iDARTS \cite{9525822} &
  97.65 &
  -- &
  75.3 &
  1.9 &
  GD &
  One-shot \\ \hline
NM(ensemble across runs) \cite{elsken2017simple} &
  95.6 &
  80.4 &
  -- &
  4 &
  RL &
  Network   morphism \\
Net2Net \cite{chen2015net2net} &
  -- &
  -- &
  78.5 &
  18 &
  RL &
  Network   morphism \\
EAS \cite{cai2018efficient} &
  95.77 &
  -- &
  -- &
  10 &
  RL &
  Network   morphism \\
Once for all \cite{cai2019once} &
  -- &
  -- &
  80 &
  1.7 &
  EA &
  Network   morphism \\ \hline
AK-DP \cite{jin2019auto} &
  96.4 &
  -- &
  -- &
  -- &
  BO &
  GP \\
NAS-BOWL   \cite{ru2020interpretable} &
  97.39 &
  -- &
  -- &
  3 &
  BO &
  GP \\
GP-NAS \cite{li2020gp} &
  96.21 &
  -- &
  73.4 &
  0.9 &
  BO &
  GP \\
BANANAS \cite{white2021bananas} &
  97.36 &
  -- &
  -- &
  -- &
  BO &
  Neural   predictor \\
BayesNAS   \cite{zhou2019bayesnas} &
  97.19 &
  -- &
  73.5 &
  0.2 &
  BO &
  One-shot \\
NAGO   \cite{ru2020neural} &
  96.6 &
  79.3 &
  76.8 &
  -- &
  BO &
  BNN \\
BONAS   \cite{shi2020bridging} &
  97.57 &
  -- &
  75.2 &
  10 &
  BO &
  GCN \\
SSA-NAS \cite{tang2020semi} &
  94.01 &
  78.64 &
  -- &
  -- &
  EA &
  GCN \\
E2EPP   \cite{sun2019surrogate} &
  94.70 &
  77.98 &
  -- &
  8.5 &
  EA &
  Random forest \\
PRE-NAS   \cite{peng2022pre} &
  97.51 &
  -- &
  76.0 &
  0.6 &
  EA &
  Random   forest \\
NPENAS   \cite{wei2020npenas} &
  97.46 &
  -- &
  -- &
  1.8 &
  EA &
  GIN+MLP \\
NSGANetV1   \cite{lu2020multi} &
  97.98 &
  85.62 &
  76.2 &
  27 &
  EA &
  Down-scale \\
NSGANetV2  \cite{lu2020nsganetv2} &
  98.4 &
  -- &
  80.4 &
  -- &
  EA &
  Ensemble \\
GATES \cite{ning2020generic} &
  97.42 &
  -- &
  -- &
  -- &
  EA &
  GATES+MLP \\
  ReNAS \cite{xu2021renas} &
  93.99 &
  78.56 &
  -- &
  -- &
  EA &
  LeNet-5 \\  \hline
\end{tabular}
\end{table*}

\section{Challenges and Future Directions}
Although a lot of effective NAS approaches have been proposed and achieved compelling performance, there are still many open challenges and future directions for efficient neural architecture search approaches.

\subsection{Sampling efficiency}

The original purpose of introducing a surrogate model as a performance predictor is to reduce the computational cost for evaluating a network (mostly training from scratch) during the searching process. However, in order to build such a surrogate with good predictive performance, it is usually necessary to construct a training set first. Building a training set itself also requires sampling and training a large number of network models by gradient-based methods. In surrogate-assisted NAS, a desired surrogate model should be not only well-performed but also sampling efficient (i.e., the number of network architectures to be trained for constructing the surrogate is as small as possible). Furthermore, the computational overhead associated with training surrogate models cannot be overlooked. For example, the computational complexity of training a Gaussian Process model is $O(N^3)$, where $N$ denotes the number of training samples. When the performance of a single surrogate is undesirable, an ensemble of different types of surrogate models needs to be build, which further increases the training cost. 

\subsection{Model management}

In offline surrogate-assisted NAS algorithms, the surrogate model is only trained once before the optimization process, while in online approaches, newly sampled networks are generated and added to the training set, and the surrogate is updated in an online manner. Considering the large size of a search space and the computational overhead for training a surrogate, it is impossible for a surrogate model to cover the entire search space. Under this circumstance, a properly designed model management strategy can improve the predictive accuracy of a surrogate. For example, in multi-objective neural architecture search, we are concerned more about the model performance near the Pareto front than those dominated regions. 

Therefore, the selection of surrogate models, the construction of the initial training dataset, and the model management strategies are key factors in surrogate-assisted NAS that deserve more attention.

\subsection{Federated Learning}

Despite the increasing application of federated learning and AutoML, only a handful of work has attempted to design effective network architectures under FL scenarios. The main challenge for training a neural network model in FL is the distributed (and even non-IID) data allocation due to privacy concerns. This becomes more notable for federated NAS since a huge amount of candidate architectures need to be trained and evaluated during the optimization process. Here are some future directions suggested for federated NAS:

\begin{itemize}

\item In the cases where data security is the primary concern for federated neural architecture search, privacy-preserving approaches such as differential privacy \cite{dwork2014algorithmic}, secure aggregation \cite{bonawitz2016practical} and homomorphic encryption \cite{acar2018survey} could be aggregated into current NAS algorithms to further enhance the security protection.

\item A significant difference between federated NAS and centralized NAS is various hardware constraints among different clients. In a centralized environment, we simply need to consider hardware limitations of the target device, and convert it into a multi-objective optimization problem. However, in federated NAS, it is more common for different participants to have various hardware constraints or even different data distribution, so it may be not reasonable to deploy a uniform global model to all clients.

\item The communication cost in federated learning is a key obstacle that hinders the algorithm from better scalability. A single network trained by FedAvg \cite{mcmahan2017communication} requires many rounds of communication to transmit the updated information between clients and server. As one can imagine, the communication overhead of an evolutionary NAS approach will increase proportionally with the population size. Therefore, it is worth investigating how to reduce the communication and computational costs of population-based NAS in a federated learning environment.

\item Finally, the effectiveness of surrogate models in federated NAS remains to be explored. For example, one method is to train a global surrogate by aggregating local information from the clients, another method is to maintain and update a local surrogate on each client separately.

\end{itemize}

\subsection{Green AI}

Generally, there is a considerable computational overhead when searching for an optimal network architecture for a given task. Although various proxy methods have been developed, it should be noted that the training process of the surrogate itself will also consume computing resources. Consequently, hardware requirements and expensive computational cost have been the bottlenecks in the real-world application of neural architecture search. It is worth investigating how to adapt existing surrogate-assisted NAS algorithms and deploy them on resource-limited edge devices, such as mobile phones, IoT devices, and embedded systems \cite{sun2022gibbon}. In fact, the majority of current NAS approaches rely on direct encoding strategies, which limits the diversity of the candidate network architectures. One promising direction is to develop indirect or generative encoding strategies with scalability, in order to enhance the flexibility for deployment on resource-constrained edge platforms.

\section{Conclusion}

\textcolor{black}{This survey conducts a systematic overview and detailed analysis of computationally efficient methods for performance prediction in neural architecture search algorithms. We first give a brief overview of existing NAS methods, mainly based on reinforcement learning and evolutionary algorithms. We categorize the computationally efficient NAS approaches into proxy-based methods and surrogate-assisted methods according to whether the weight values are needed for the prediction. The proxy-based methods evaluate the network performance using proxy metrics, including low-fidelity estimation, one-shot NAS and network morphism. We present a summary of representative literature on each type of strategy with an analysis of their characteristics. In contrast to other NAS surveys, we further concentrate on surrogate-assisted NAS methods, where surrogate models are trained and updated to evaluate the performance of unseen architectures in an end-to-end manner.} Then, a detailed  description and performance analysis of different types of surrogate models with corresponding milestone work. Finally, we discuss the existing challenges and future directions for performance prediction in NAS optimization, especially under the privacy-preserving federated learning framework. We hope this survey will provide some insights into more efficient performance prediction in NAS algorithms, thereby promoting research in AutoML and security machine learning.


%

\ifCLASSOPTIONcaptionsoff
  \newpage
\fi





\bibliographystyle{IEEEtran}
\bibliography{IEEEexample.bib}
\end{document}